\documentclass[10pt,twocolumn,letterpaper]{article}

\usepackage{iccv}
\usepackage{times}
\usepackage{epsfig}
\usepackage{graphicx}
\usepackage{amsmath}
\usepackage{amssymb}
 \usepackage{graphicx}
 \usepackage[table,xcdraw]{xcolor}
\usepackage[accsupp]{axessibility} 

\usepackage[bookmarks=false,colorlinks,bookmarks=false]{hyperref}

\iccvfinalcopy 

\def\iccvPaperID{12601} 

\ificcvfinal\fi

\begin{document}

\title{Multi-label affordance mapping from egocentric vision}


\author{Lorenzo Mur-Labadia \qquad Jose J. Guerrero \qquad Ruben Martinez-Cantin\\
I3A - Universidad de Zaragoza\\
{\tt\small {lmur, jguerrer, rmcantin}@unizar.es}
}



\maketitle
\ificcvfinal\thispagestyle{empty}\fi

\begin{abstract}

Accurate affordance detection and segmentation with pixel precision is an important piece in many complex systems based on interactions, such as robots and assitive devices. We present a new approach to affordance perception which enables accurate multi-label segmentation. Our approach can be used to automatically extract grounded affordances from first person videos of interactions using a 3D map of the environment providing pixel level precision for the affordance location. We use this method to build the largest and most complete dataset on affordances based on the EPIC-Kitchen dataset, \href{https://github.com/lmur98/epic_kitchens_affordances}{\textcolor{blue}{EPIC-Aff}}, which provides interaction-grounded, multi-label, metric and spatial affordance annotations. Then, we propose a new approach to affordance segmentation based on multi-label detection which enables multiple affordances to co-exists in the same space, for example if they are associated with the same object. We present several strategies of multi-label detection using several segmentation architectures. The experimental results highlight the importance of the multi-label detection. Finally, we show how our metric representation can be exploited for build a map of interaction hotspots in spatial action-centric zones and use that representation to perform a task-oriented navigation.
\end{abstract}

\section{Introduction}

When humans repeatedly interact in a close environment, we associate a set of affordable actions with a certain distribution of objects in a physical space. For example, we associate a pan on a stove with cooking, but the same pan on the sink with washing. A joined spatial-semantic understanding contains powerful insights to understand human behaviour. 
This requires a close combination of perception, mapping and navigation algorithms; with potential applications in augmented reality systems \cite{quesada2022holo, quesada2022proactive}, but also guiding a robot \cite{do2018affordancenet, koppula2015anticipating} or assistive devices \cite{sanchez2020semantic}. In the last years, the ability of deep learning models to extract high-level representations has improved the perception of autonomous agents, while egocentric vision offers a powerful viewpoint for modelling human-object interaction understanding. Recent advances include anticipating future actions \cite{furnari2019would, girdhar2021anticipative, abu2018will}, model the hands-object manipulation \cite{furnari2017next, zhang2022fine, goyal2022human, damen2016you}, detect the change in an object state \cite{alayrac2017joint}, identify interaction hotspots \cite{fang2018demo2vec, nagarajan2019grounded} or create topological maps \cite{nagarajan2020ego}. Despite the fast movements of a headset camera, egocentric perception has also contributed to the mapping and planning phases: localising the agent in a known 3D map \cite{liu2022egocentric}, performing visual navigation \cite{nagarajan2022egocentric, ramakrishnan2021environment} or building \emph{third-person} (allocentric) maps \cite{cartillier2021semantic, maturana2018real}.



\begin{figure}[t]
\centering
\includegraphics[width=0.99\columnwidth]{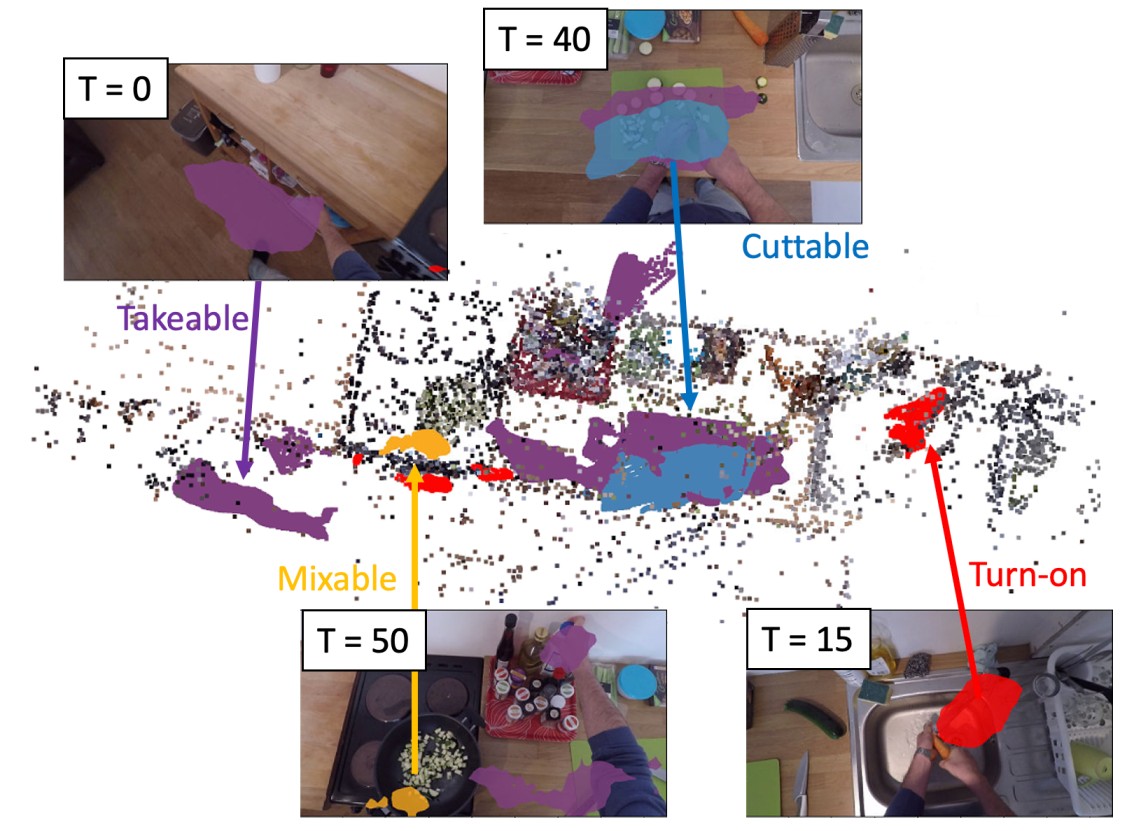}
\caption{From a sequence of egocentric observations, our agent creates a spatial-metric multi-label representation of the affordances, enabling a task-oriented navigation.}
\label{fig:teaser_fig}
\end{figure}

Gibson's perception theory presents affordances as the potential actions that the environment offers to the agent based on its motor capabilities \cite{gibson1977theory}. For example, the person can afford \textit{taking} a glass, but the affordances of a soup in a pan can be \textit{mixing, emptying, scooping} and \textit{pouring} simultaneously. This multiplicity models better complex dynamic environments and opens the door to multi-agent collaboration with task synchronization. Although some authors have focused on more complex affordance models \cite{montesano2008learning, yu2023fine}, affordance perception is typically defined as a classification problem. Some authors have focused instead on grounded affordances \cite{fang2018demo2vec}, which provide a more flexible setup and are truly associated with motor capabilities, showing improvements in action anticipation \cite{luo2022learning}. However, most learning approaches in affordance perception consider the problem ungrounded to the agent interaction with the object, requiring previous annotations of each affordance occurrence \cite{myers2015affordance, nguyen2017object, minh2020learning, do2018affordancenet, caselles2021standard, nguyen2016detecting}. While ungrounded methods have the advantage of providing pixel-wise precision, which we call metric understanding of the scene, many grounded approaches rely on full image classification losing any metric meaning. In this paper, we propose a grounded approach with pixel-wise precision, which enables detailed metric understanding while maintaining the flexibility of grounded methods and that can be used as prior information for more complex affordance models \cite{yu2023fine}. Close to our proposal is the work of Nagarajan et al. \cite{nagarajan2019grounded}, which presented a grounded approach for extracting interaction hotspots by directly observing videos. Similar to other previous works, the hotspots are modelled as a single available affordance. Instead, we propose to consider the multiplicity of affordances for a single object or spatial zone through multi-label pixel-wise predictions.



We build a pipeline to automatically collect multi-label pixel-wise annotations from real-world interactions using a temporal, spatial and semantic representation of the environment. We use this method with the EPIC Kitchens videos \cite{damen2018scaling} to build a dataset of grounded affordances (EPIC-Aff), which to the author's knowledge, constitute the largest dataset in affordance segmentation up to date. We then adapt several segmentation architectures to the multi-label paradigm to extract more diverse information from the scene based on the assumption that the same object may have multiple affordances available.  Using a mapping approach 
we extract the multi-label affordance segmentation to build a map that spatially links activity-centric zones as shows Figure \ref{fig:teaser_fig}, allowing a metric representation of the environmental affordances and goal-directed navigation tasks. Finally, we perform a quantitative evaluation of the extension from common architectures to the multi-label paradigm and we show mapping and planning applications of our approach, that can be used for assistive devices or in robotic scenarios.

\section{Related works}

\subsection{Learning Visual Affordances}

Ungrounded approaches \cite{myers2015affordance, nguyen2017object, minh2020learning, do2018affordancenet, caselles2021standard, nguyen2016detecting} for affordance perception are fully supervised by manually labelled masks. Due to their similarity with semantic segmentation or object detection tasks, these works use common architectures such as an encoder-decoder \cite{nguyen2016detecting}, proposal-based detectors \cite{do2018affordancenet, nguyen2017object, minh2020learning, caselles2021standard} or Bayesian instance segmentation \cite{aff_loren}. Concerning grounded works, Fang et al. \cite{fang2018demo2vec} extracted a latent representation from demonstration videos or Luo et al. \cite{luo2022learning} transferred the learning from exocentric images to the egocentric perspective using only the semantic label as supervision. Nagarajan et al. \cite{nagarajan2019grounded} extracted the interaction hotspots by deriving the gradient-weighted attention maps obtained at training an action classifier on videos. Then Ego-Topo \cite{nagarajan2020ego} built a topological graph of the scene to perform affordance classification from egocentric videos, grouping each node visually and temporally coherent frames with similar object and action distributions. This allows them to discover activity-centric zones based on their visual content and represent semantically the traversed paths with the edges of the graph. 

\subsection{Multi-label perception}

In the multi-label segmentation task, we assign two or more categories to a single pixel. A particular case is an amodal segmentation where the relevance of occluded parts depends on the depth order \cite{zhu2017semantic, mohan2022amodal}. The most common applications of multi-label segmentation are biomedical works, where there are multiple overlapped non-exclusive levels of tissues. Existing architectures extend a U-Net with minor changes such as a dynamic segmentation head \cite{deng2022single}, shuffle-attention mechanisms in the skip-connections \cite{lempart2022pelvic}, a combination of appearance and pose features \cite{bonheur2019matwo} and split-attention modules \cite{lee2020split}. 
The closest approach to multi-label segmentation is multi-label image recognition, where the label imbalance between positives and negatives in each binary classifier and the extraction of features from multiple objects make this task more complex \cite{liu2021query2label}. Class distribution aware losses \cite{wu2020distribution} such as the asymmetric loss \cite{ben2020asymmetric}, the focal loss \cite{lin2017focal} or the Multi-Label Softmax loss \cite{ge2018chest} correct the over-suppression of negative samples. On the other hand, Graph Neural Networks such as \cite{ye2020attention} deal with the feature extraction from multiple objects by creating a dynamic graph for each image that leverages the content-aware category representations. Finally, the transformer architecture extracts multiple attention maps in the different regions of interest \cite{cheng2021mltr, lanchantin2021general}, guiding the multi-label classification \cite{liu2021query2label} or ranking the class of the pixels considering only the categories selected by the classifier \cite{he2022rankseg}.

\section{Grounded Affordance Labelling}

\begin{figure}
\centering

\includegraphics[width=0.99\columnwidth]{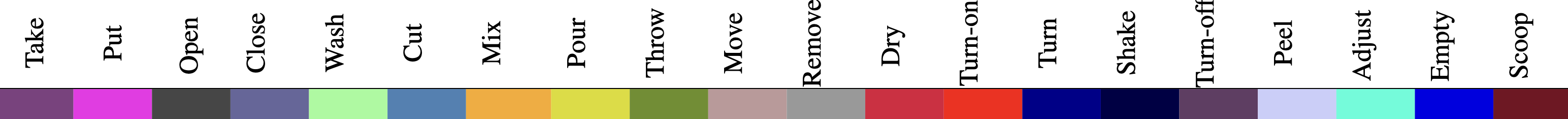} \\ 
\includegraphics[width=0.32\columnwidth]{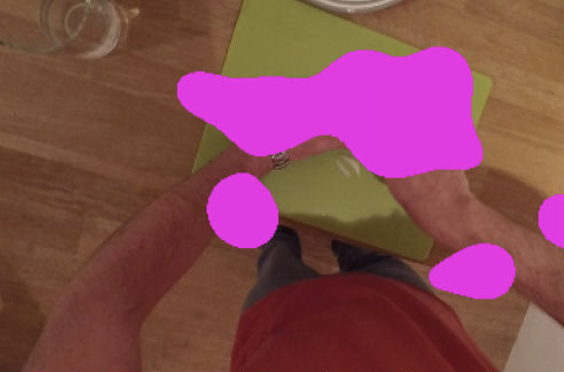}  
\includegraphics[width=0.32\columnwidth]{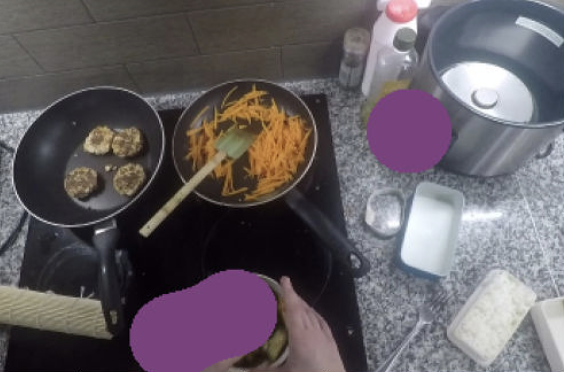} 
\includegraphics[width=0.32\columnwidth]{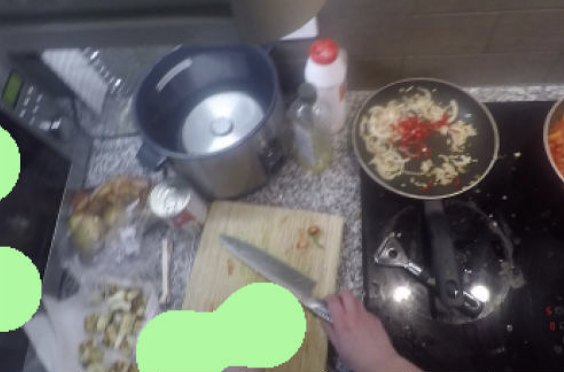}  \\

\includegraphics[width=0.32\columnwidth]{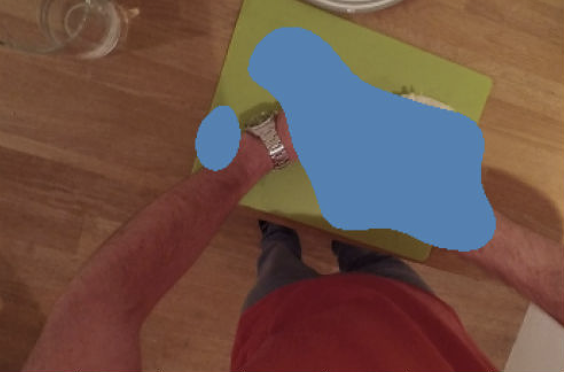}  
\includegraphics[width=0.32\columnwidth]{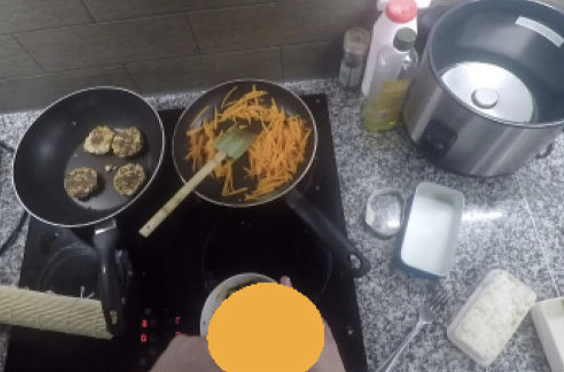} 
\includegraphics[width=0.32\columnwidth]{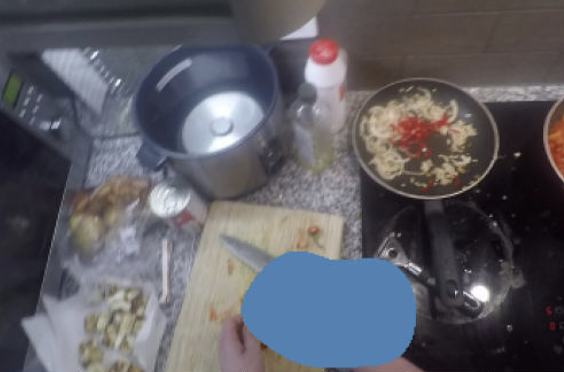} \\

\includegraphics[width=0.32\columnwidth]{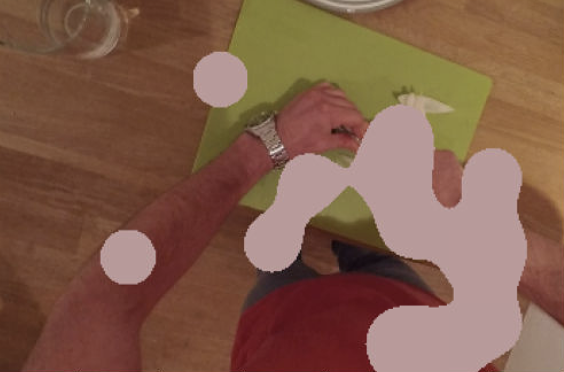}  
\includegraphics[width=0.32\columnwidth]{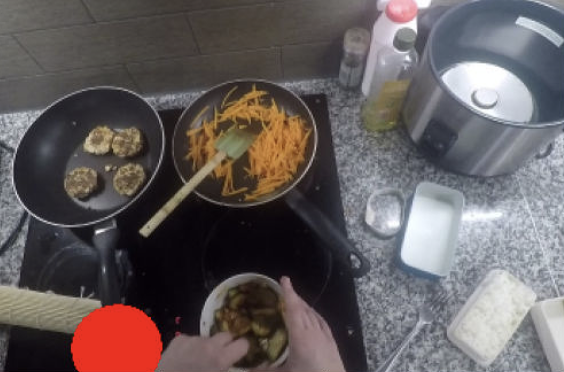} 
\includegraphics[width=0.32\columnwidth]{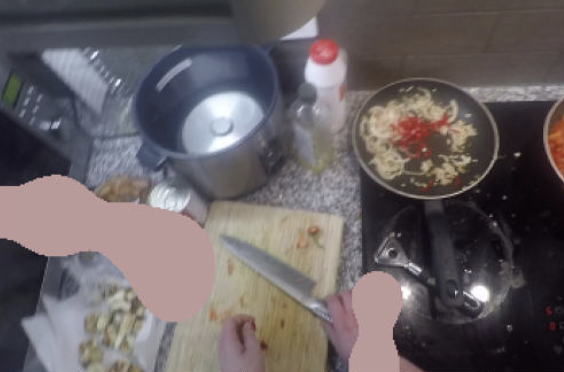}  \\

\includegraphics[width=0.32\columnwidth]{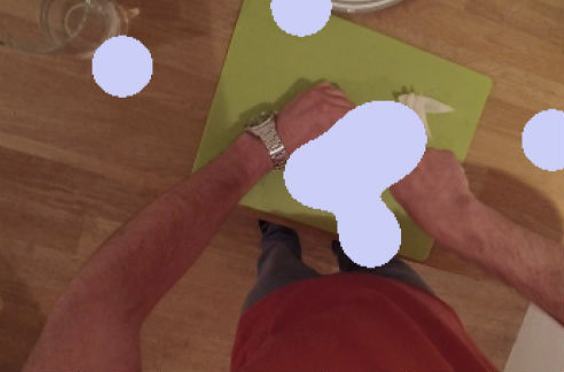}  
\includegraphics[width=0.32\columnwidth]{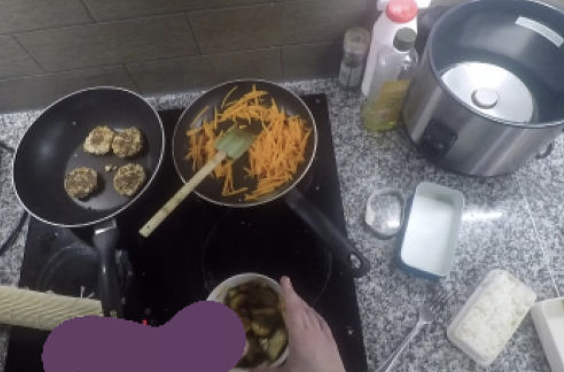} 
\includegraphics[width=0.32\columnwidth]{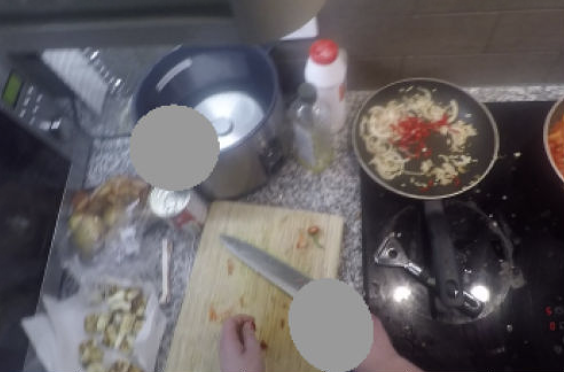}  \\

\caption{Ground truth examples. For visualization purposes, we show a single label of the affordable action on its location, although these are overlapped for the same sample. The food in the bowl affords \textit{taking} or \textit{mixing}, while the cutting board on the left affords \textit{putting}, \textit{cutting}, \textit{moving} and \textit{peeling}.}
\label{fig:dataset_examples}
\end{figure}

We extract \textit{automatic}, \textit{interaction-grounded}, \textit{multi-label pixel-wise} and \textit{spatial} affordance annotations from a sequence of real-world images in complex and cluttered environments, as shows Figure \ref{fig:dataset_examples}. Our multi-label segmentation approach learns \textit{all} the potential options and does not reduce the perception to a single action. For example, a potato on a chopping board offers \textit{cutting, putting, peeling, removing} and \textit{taking} simultaneously. Current affordance segmentation works \cite{do2018affordancenet, myers2015affordance, nguyen2017object, luo2022learning, aff_loren} assume a single-label affordance per object and lose a valuable amount of information. Although other affordance models allow for multiple predictions, these works ignore the segmentation of the interaction hotspot in the image and lose the pixel-level accuracy of the segmentation models. For example, topological maps extract multiple affordances from an image \cite{nagarajan2020ego}, or action anticipation models predict a probability distribution of the different possibilities \cite{furnari2019would, girdhar2021anticipative}. Our methodology gets the best of two worlds producing multi-label metric masks, resulting in a full distribution of affordances. It enables a deeper understanding of the manipulation task such as the grasping points of the tool \cite{ardon2019learning} or the evolution of the manipulation process over time\cite{liu2022joint}. Similar to previous unsupervised or weakly supervised methods \cite{montesano2008learning}, we extract affordance labels from weak VISOR and EPIC Kitchens annotations grounded on actual interactions.

We join the affordances with their 3D spatial location by extracting the camera poses. The spatial approach to affordance perception is not new for the community. Rhinehart et al. \cite{rhinehart2016learning} associate the functionality of regions with specific spatial locations, showing that that defining an affordance based solely on semantics is insufficient due to the significant influence of the physical context. For example, a frying pan is only \textit{cookable} when it is on the hob or a plate is \textit{washable} when the agent is next to the dishwasher. However, their method results in smooth 2D maps, which can be problematic for the fine-grained affordances in 3D space in our EPIC-Aff dataset. Instead, our method is able to scale up to large environments while maintaining the detail by using neural networks. Other previous works \cite{guan2020generative, rhinehart2017first} also use SLAM for action prediction but with addressing different problems. In those works, the action is set on the human, while the image provides context; while in our case the action/affordance is set on the environment and the user provides context. 
In our work, we use COLMAP to extract the relative pose between sparse frames with a filter of the dynamic objects, registering up to 93 $\%$ of the frames compared with the 44 $\%$ of the frames registered by ORB-SLAM \cite{mur2015orb} on EPIC Kitchens \cite{nagarajan2020ego}. Recently, EPIC Fields \cite{tschernezki2023epic} registered the camera pose of the dense videos in EPIC Kitchens using neural rendering techniques.


\begin{table}
\begin{center}
\resizebox{\columnwidth}{!}{
\begin{tabular}{c|c|cccc|ccc}
\hline
Dataset & Year & IG & Pix & ML & CP & \#Obj. & \#Aff. & \#Imgs. \\ \hline
UMD \cite{myers2015affordance} & 2017 & \text{\sffamily X} & \checkmark & \text{\sffamily X} & \text{\sffamily X} & 17 & 7 & 30,000 \\
IIT-Aff \cite{nguyen2017object} & 2017 & \text{\sffamily X} & \checkmark &\checkmark & \text{\sffamily X} & 10 & 9 & 8,835 \\
ADE-Aff \cite{chuang2018learning} & 2018 & \text{\sffamily X} & \checkmark & \checkmark & \text{\sffamily X} & 150 & 3 & 10,000 \\
OPRA \cite{fang2018demo2vec}& 2018 & \checkmark &\checkmark & \text{\sffamily X} & \text{\sffamily X} & - & 7 & 20,774 \\
Grounded I.H \cite{nagarajan2019grounded}& 2018 & \checkmark & \checkmark & \text{\sffamily X} & \text{\sffamily X} & 31 & 20 & 1,800$^{*}$ \\
Ego-Topo \cite{nagarajan2020ego} & 2020 & \checkmark & \text{\sffamily X} & \checkmark & \text{\sffamily X} & 304 & 75-120 & 1,020-1,115$^{*}$ \\
PAD v2 \cite{zhai2022one} & 2021 & \text{\sffamily X} & \checkmark & \text{\sffamily X} & \text{\sffamily X} & 72 & 31 & 30,000 \\
AGD20k \cite{luo2022learning} & 2022 & \text{\sffamily X} & \checkmark & \text{\sffamily X} & \text{\sffamily X} & 50 & 36 & 23,816 \\ \hline
EPIC-Aff & 2023 & \checkmark & \checkmark & \checkmark & \checkmark & 304 & 20-43 & 38,876 \\ \hline
\end{tabular}
}
\end{center}
\caption{\textbf{Visual affordance datasets statistics}. I.G: Interaction Grounded. Pix: pixel-wise annotations. ML: multi-label. CP: camera poses \#Obj: Number of objects. \#Aff: Number of affordances. \#Imgs: total number of images. $^{*}$ The affordance labels are only for evaluation, the model is trained supervised only by action labels provided by \cite{damen2018scaling}}
\label{tab:datasets}
\end{table}


\subsection{Affordance datasets}

Following our motivation, we conduct a study along the visual affordance datasets shown in Table \ref{tab:datasets}. The ungrounded datasets are subjected to the annotator's consideration and required to draw pixel-wise semantic mask to each object part \cite{nguyen2017object, myers2015affordance, chuang2018learning, zhai2022one, luo2022learning} or additional sensors \cite{koppula2014physically, castellini2011using}, decreasing the object variability and limiting the scalability due to the annotation costs. The UMD dataset contains a semantic affordance map for objects in isolated conditions and with low variability,  which prevents generalisation \cite{myers2015affordance}. The IIT-Aff dataset \cite{nguyen2017object} provides the most comprehensive annotations designed for use in robotics, including multiple objects in a single image. The ADE-Aff dataset \cite{chuang2018learning}, built on top of ADE20K scenes, examines the social acceptability of actions about context but is limited to only three affordance classes. The AGD20k \cite{luo2022learning} dataset includes the largest number of categories and actions by transferring from an exocentric to an egocentric viewpoint perspective. 
On the other hand, grounded works learn from observing interactions on the EPIC-Kitchens sequences \cite{damen2018scaling}, internet demonstration videos \cite{fang2018demo2vec} or with gaze point with eye-tracking devices \cite{fathi2012learning}. The annotations provided are only used for evaluation since they do not require strong supervision. However, these approaches ignore the pixel-wise precision \cite{nagarajan2020ego} or the multi-label modality of our approach\cite{nagarajan2019grounded}.

Based on the mentioned limitations, our novel dataset EPIC-Aff provides multi-label pixel-wise affordance annotations with the camera pose. It contributes to a diverse and comprehensive affordance database with the largest number of images up to date. This better captures the complexity, dynamics, multiplicity and variability of real-world environments, such as preparing a recipe in a kitchen. Finally, as our labels are automatically extracted, we enable the application of our method to other egocentric datasets.

\begin{figure*}
\centering
\includegraphics[width=0.99\textwidth]{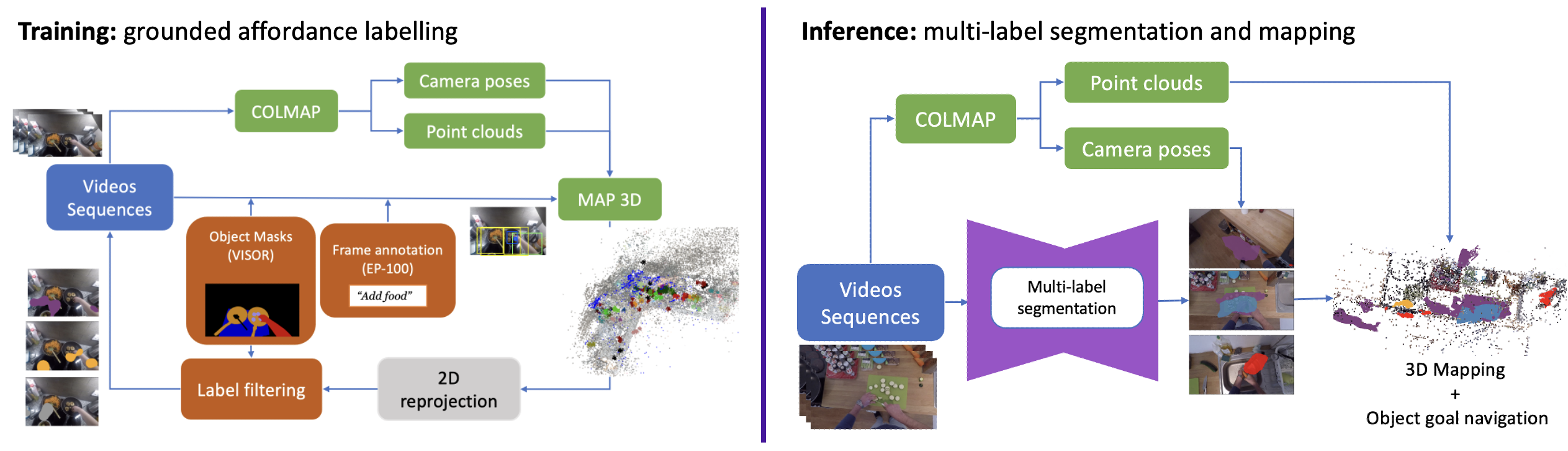}  \\
\caption{Left: Pipeline with the automatic extraction of the pixel-wise labels on the EPIC-Aff dataset. We combine the EPIC-100 narration with the VISOR masks annotations to extract the interaction point. Then, using the camera pose extracted from COLMAP, we project all the interactions in a common 3D global reference. Finally, we reproject all the past interactions to each frame, and filter the affordance annotation by the objects present at the image. Right: the multi-label masks predictions from our model are leveraged to a 3D map}
\label{fig:main_img}
\end{figure*}

\subsection{EPIC-Aff dataset}

We detail the procedure shown in Figure \ref{fig:main_img} for our grounded affordance labelling. EPIC-Aff\footnote{\url{https://github.com/lmur98/epic_kitchens_affordances}} is composed of 38,876 images with up to 43 different affordable actions $\mathcal{K}$. We choose the EPIC-Kitchens as the base dataset because of its sequential and repetitive nature, which allows us to extract the 3D geometry, and because the kitchen is a scenario with multi-step and structured activities very rich in semantics. We cover all the object categories present in the EPIC-100 annotations, which constitute a wide, large and diverse knowledge base.

\begin{figure}[htbp]
\centering
\includegraphics[width=0.49\columnwidth]{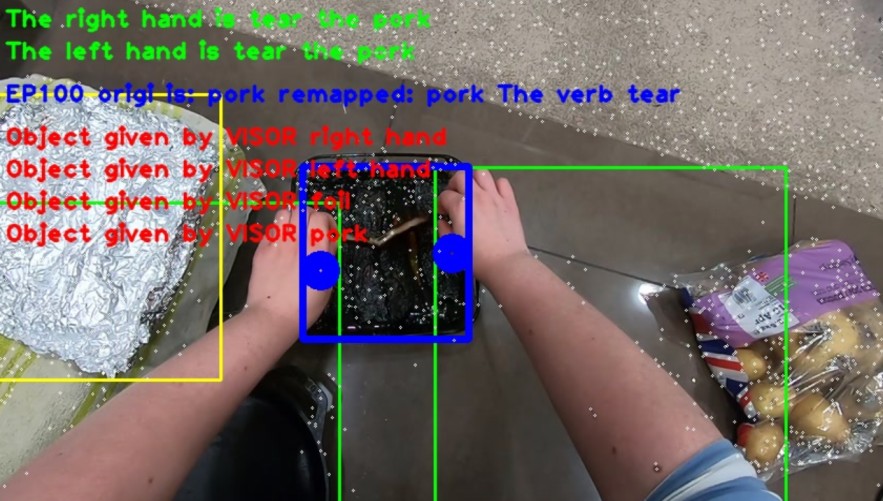} 
\includegraphics[width=0.49\columnwidth]{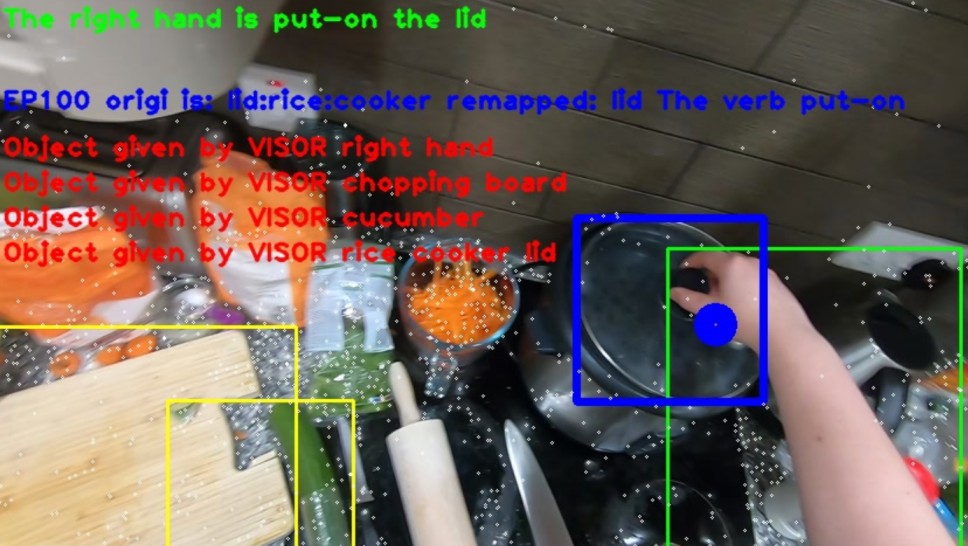} \\
\caption{Using the masks provided by VISOR Kitchens, we define the intersection between the object and the hand bounding boxes as the center of the interaction. We show in yellow the bounding box of the non-interacting objects, in green the bounding box of the hands and in blue the bounding box of the interacting object.}
\label{fig:interactiong_examples}
\end{figure}

From a sequence of video, we join the EPIC-100 narrations \cite{damen2018scaling} and the VISOR Kitchen annotations \cite{darkhalil2022epic} to obtain a sparse sequence of frames $\mathcal{S_M}= (f_1, ..., f_N)$ with the localization of the interactions on the image, as shows Figure \ref{fig:interactiong_examples}. 
The EPIC-100 labels \cite{damen2018scaling} contains narrations formed by an action verb $\mathcal{V}$ with an associated object $\mathcal{O}$, i.e: "add steak", for more than 100 hours of video. VISOR Kitchens \cite{darkhalil2022epic} interpolates from sparse annotations to generate semantic masks $\mathcal{M}$ and bounding boxes $\mathcal{B}$ on the active objects. We set the centre of the interaction $x_i = \{u_i, v_i\}$ in the middle of the intersection between the hand $\mathcal{B}_h$ and the interacted object $\mathcal{B}_O$ bounding-boxes given by the narration $\mathcal{V} + \mathcal{O}$.

Then, we apply COLMAP, a Structure-from-Motion (SfM) algorithm  \cite{schoenberger2016mvs} to obtain the camera poses $T_w^c$ and a point cloud of the environment $\{X_p\}$. In the EPIC-Kitchens \cite{damen2018scaling}, each kitchen is composed of multiple videos, thus, we join all the sparse frames with interactions $\mathcal{S}_K = \{\mathcal{S}_1\ldots \mathcal{S}_M\}$ to relate frames from different videos in a common 3D reference. Furthermore, we use the VISOR semantic masks to avoid including dynamic objects in the point cloud.
    
Next, we employ a robust depth estimator based on a neural network $d_{NN} = f_d(\cdot)$ \cite{Ranftl2022} to predict the depth of interaction points $d_{NN}(x_i)$. Because the neural network computes the depth up-to-scale, we compute a scale correction factor per image to fit the network scale to the SfM scale: $scale = median(d^{SfM}(X_p) / median (d^{NN}(X_p))$ \cite{klodt2018supervising}, where $d^{SfM}(X_p)$ is the depth of all the points $\{X_p\}$ visible from the current image and $d^{NN}(X_p)$ is the depth of the same points given by the network estimator.  
    
\begin{figure} [htbp]
\centering
\includegraphics[width=0.97\columnwidth]{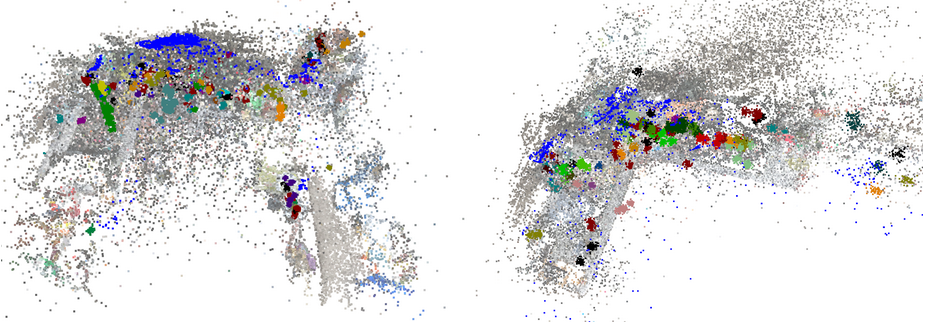} 
\caption{Historical with all the past interaction frames in that environment, where the blue dots represent the camera poses of the sparse frames from all the sequences}
\label{fig:historic maps}
\end{figure}

Using the predicted depth and the scale projection, we can project the interaction point $x_i$ in 3D space $X_i$ and use the camera pose to project into the global coordinates $X^{w}_i = T_w^c \cdot X^{c}_i$. At this point, as shown in Figure \ref{fig:historic maps}  we obtain in a common reference a history of all the interactions that occured in the kitchen $\mathcal{I}_k = \{X^{w}_1, X^{w}_2, \ldots, X^{w}_k\}$, cross-generalizing for the different sequences. This constitutes our knowledge base that follows our hypothesis that the distribution of affordances is spatially linked to pre-determined physical spaces (\textit{i.e} you only wash in the dishwasher), not only to the semantic context of a topological graph \cite{nagarajan2020ego}.

Then, once we store all the past interactions $\mathcal{I}_k$ with their $\mathcal{V}_k$ and  $\mathcal{O}_k$ labels, we reproject them back to the new camera reference system $X^{c}_i = T_c^w \cdot X^{w}_i$. Instead of considering all the object semantic masks as the affordance region, we centre a Gaussian distribution over each affordance re-projected point $X^{c}_i$ and build an additive heatmap. Then, the affordance masks $\mathcal{M}^{aff}_i$ are defined as the regions where the heatmap is greater than 0.25. This is grounded in how humans interact with objects \cite{nagarajan2019grounded} and allows us to consider the different affordability of the object parts (a knife is only \textit{graspable} with the handle). In order to generate the affordance labels $\mathcal{A} = \{ (\mathcal{V}_1, \mathcal{O}_1,  \mathcal{M}^{aff}_1), ...., (\mathcal{V}_j, \mathcal{O}_j) \}$, we select only those verbs whose associated object  $\mathcal{O}_i$ was present in the VISOR annotations $\{ \mathcal{M}, \mathcal{B} \}$. With this procedure,  we are grounding our dataset in the past interactions in that environment and associating multiple affordances to a single object. We show qualitative samples of the EPIC-Aff in Figure \ref{fig:dataset_examples}.

\begin{figure}[htbp]
\centering
\includegraphics[width=\columnwidth]{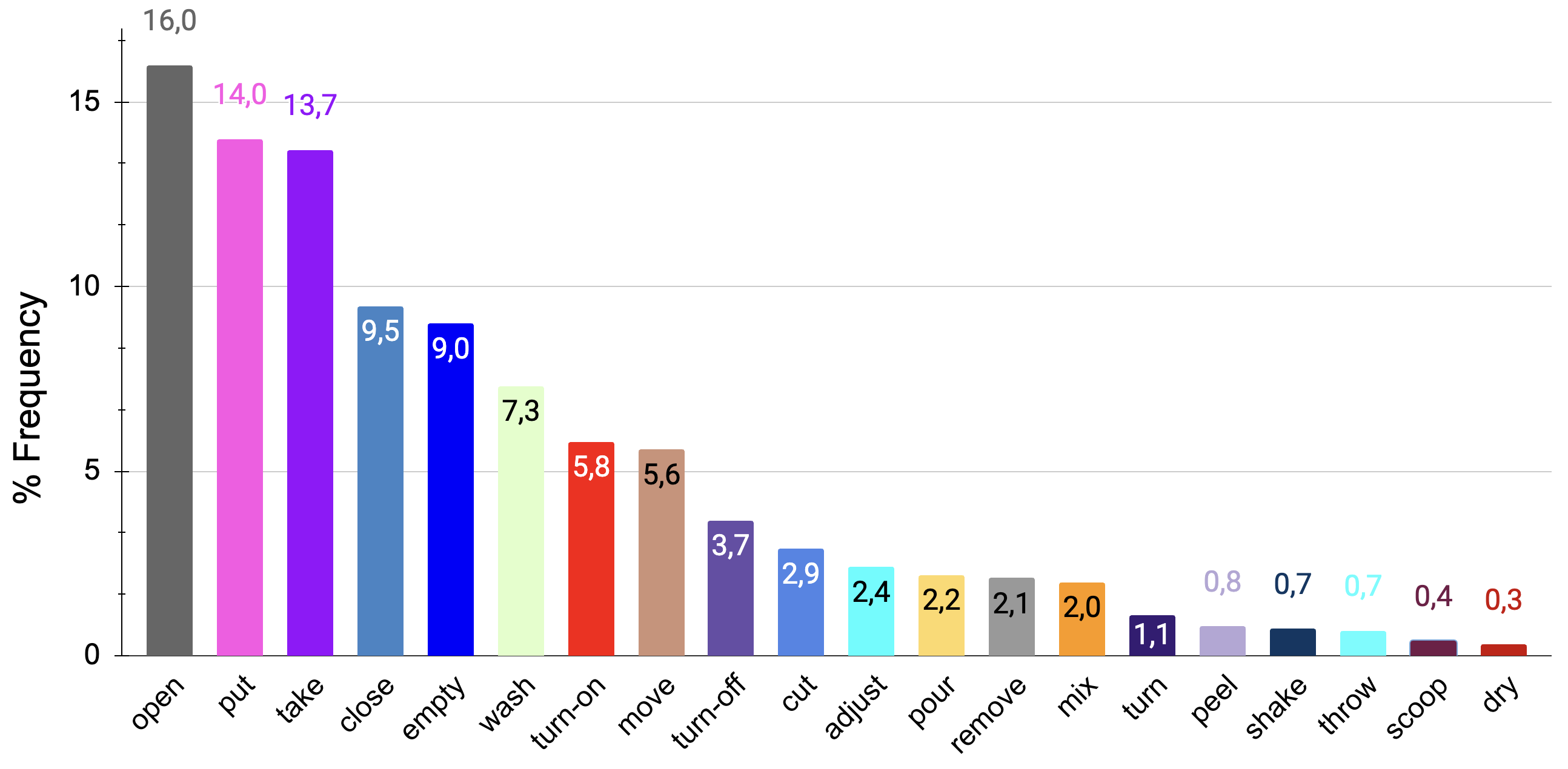}
\caption{Distribution of the 20 classes in the easy-EPIC Aff dataset, showing a significant class imbalance.}
\label{fig:distrib_classes}
\end{figure}

\begin{figure}[htbp]
\centering
\includegraphics[width=0.72\columnwidth]{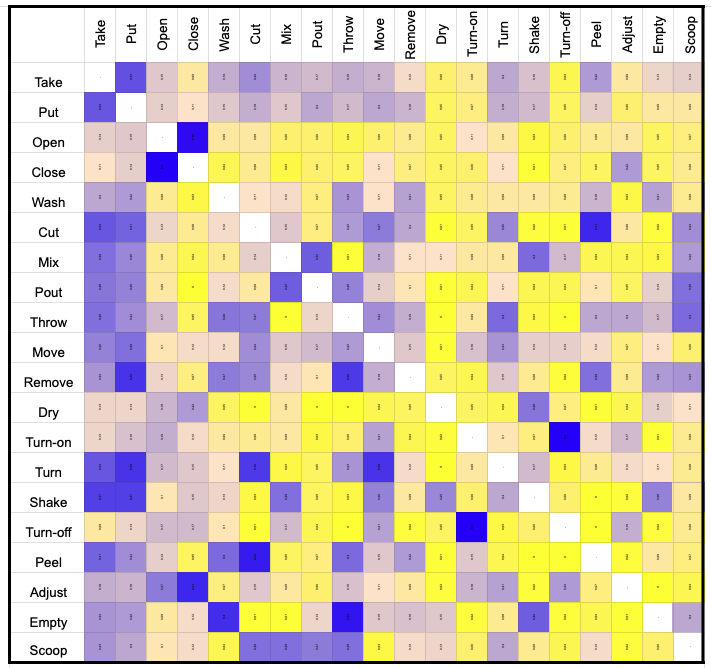}\\
\caption{Pixel ratio of the 20 classes in the easy-EPIC Aff dataset, where \textit{blue} represents high correlation between the two classes and \textit{yellow} that they do not used to concur in the same pixel.}
\label{fig:distrib_classes_2}
\end{figure}

We provide two different versions of the dataset: the easy EPIC-Aff and the complex-EPIC Aff, with 20 and 43 affordance classes respectively. There is a challenging class imbalance, as shown the Figure \ref{fig:distrib_classes} with a significant frequency gap between the most common class (\textit{open}, with a 16.0 $\%$) and the less represented (\textit{dry}, with a 0.3 $\%$). In Fig. \ref{fig:distrib_classes_2} we show the pixel ratio, which reflects that semantically similar or opposite actions are associated with the same space (\textit{i.e, turn-on, turn-off, adjust} or \textit{cut} and \textit{peel}) and the importance of the multi-level approach. This shows that activity-centric zones are physical spaces where there occur multiple common activities, both synonyms or antonyms. For example, in the hob controls region we likely find (\textit{i.e, turn-on, turn-off, adjust)}, while on the dishwasher zone, we will encounter \textit{wash} and \textit{dry}.

\section{Multi-label segmentation and mapping}

In this section, we explain our inference procedure. First, we describe the modifications needed to obtain a multi-label segmentation model. We then show how our approach can be applied to mapping and planning tasks.

\subsection{Multi-label segmentation}

In this section we describe how we transform classical semantic segmentation models to a multi-label version. While there exists lots of single-label segmentation \cite{chen2022vision, lin2016efficient, badrinarayanan2017segnet, he2017mask, he2022rankseg, chen2017deeplab} and multi-label image classification works \cite{wu2020distribution, ge2018chest, chen2019multi, liu2021query2label}, the multi-label segmentation is a more unexplored task restricted to small domains like biomedical images \cite{lempart2022pelvic, bonheur2019matwo}.

Given an input image $\mathbf{X}$, the multi-label segmentation goal is to predict a group of categories for each pixel. Therefore, we assume that each pixel could represent multiple affordances (\emph{takeable}, \emph{cuttable}, \emph{washable}, \ldots, etc.) or not belong to any category. For a total number of $\mathcal{K}$ classes we define the label $y$ for each pixel of the image as $y = [ y_1, ..., y_{k}]$, where $y_k = 1$ if the pixel contains the $\mathcal{K}$-category label, otherwise $y_k = 0$. In order to predict multi-label segmentation masks, we have evaluated two different approaches. First, we use a standard multiclass segmentation networks and evaluate three different heuristics to select multiple labels per pixel. Then, we modify the segmentation networks to output multiple binary classifiers which enable multiple labels to be active.

\begin{figure}[htbp]
\centering
\includegraphics[width=0.3\columnwidth]{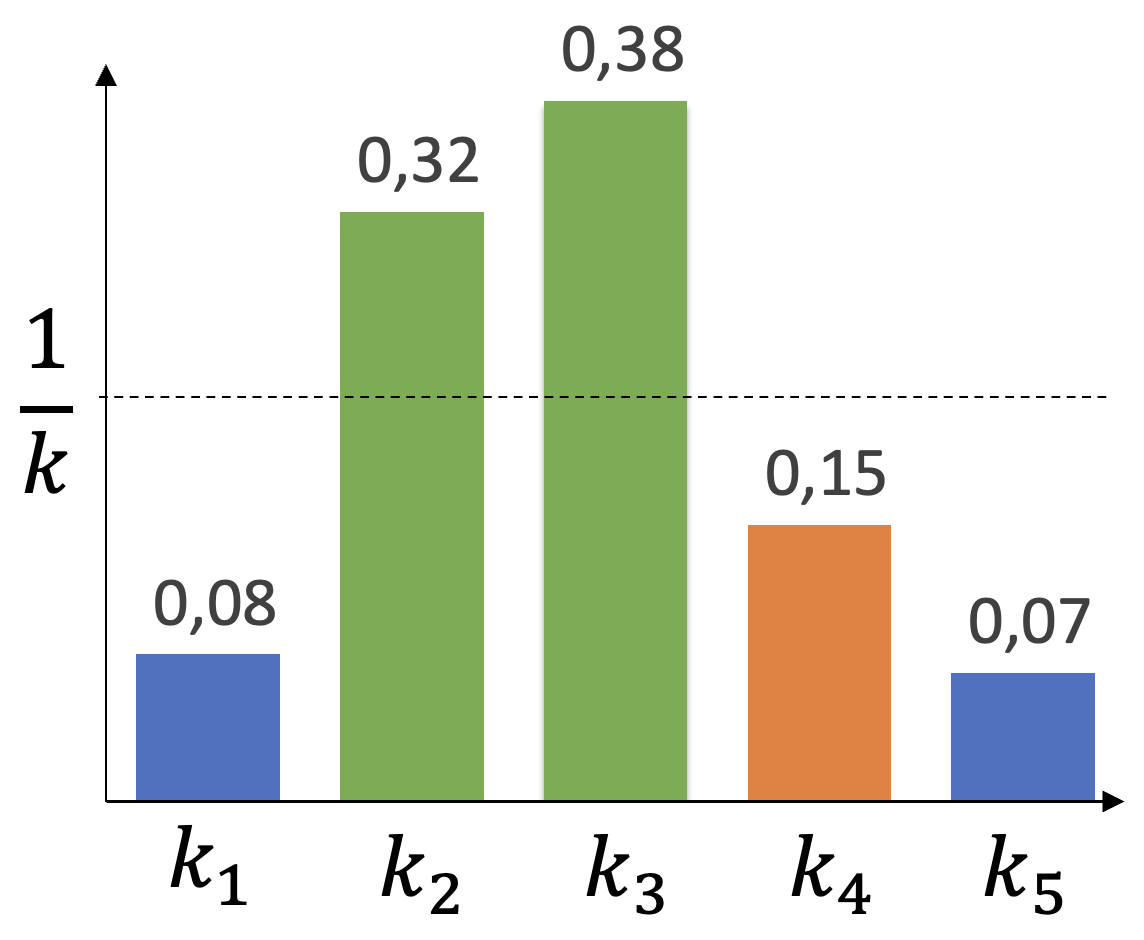} 
\includegraphics[width=0.3\columnwidth]{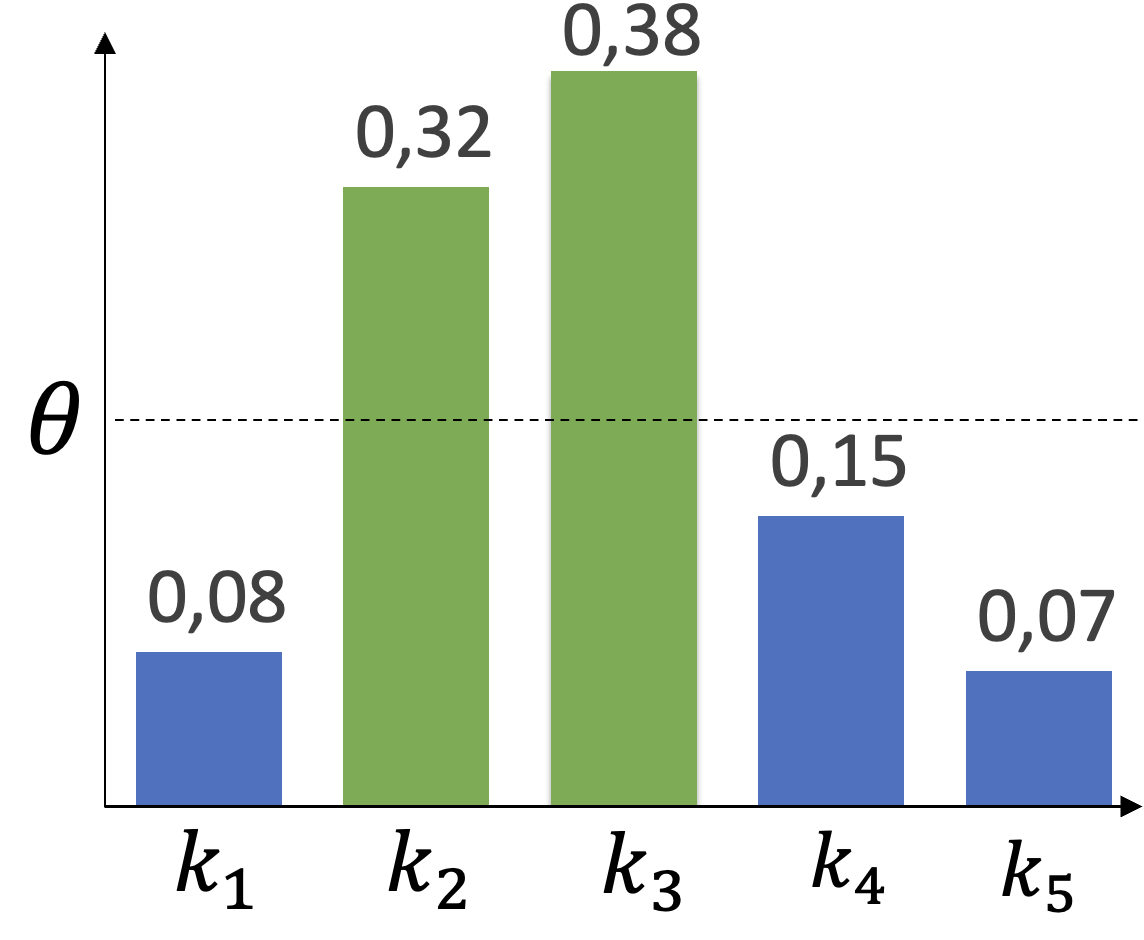} 
\includegraphics[width=0.34\columnwidth]{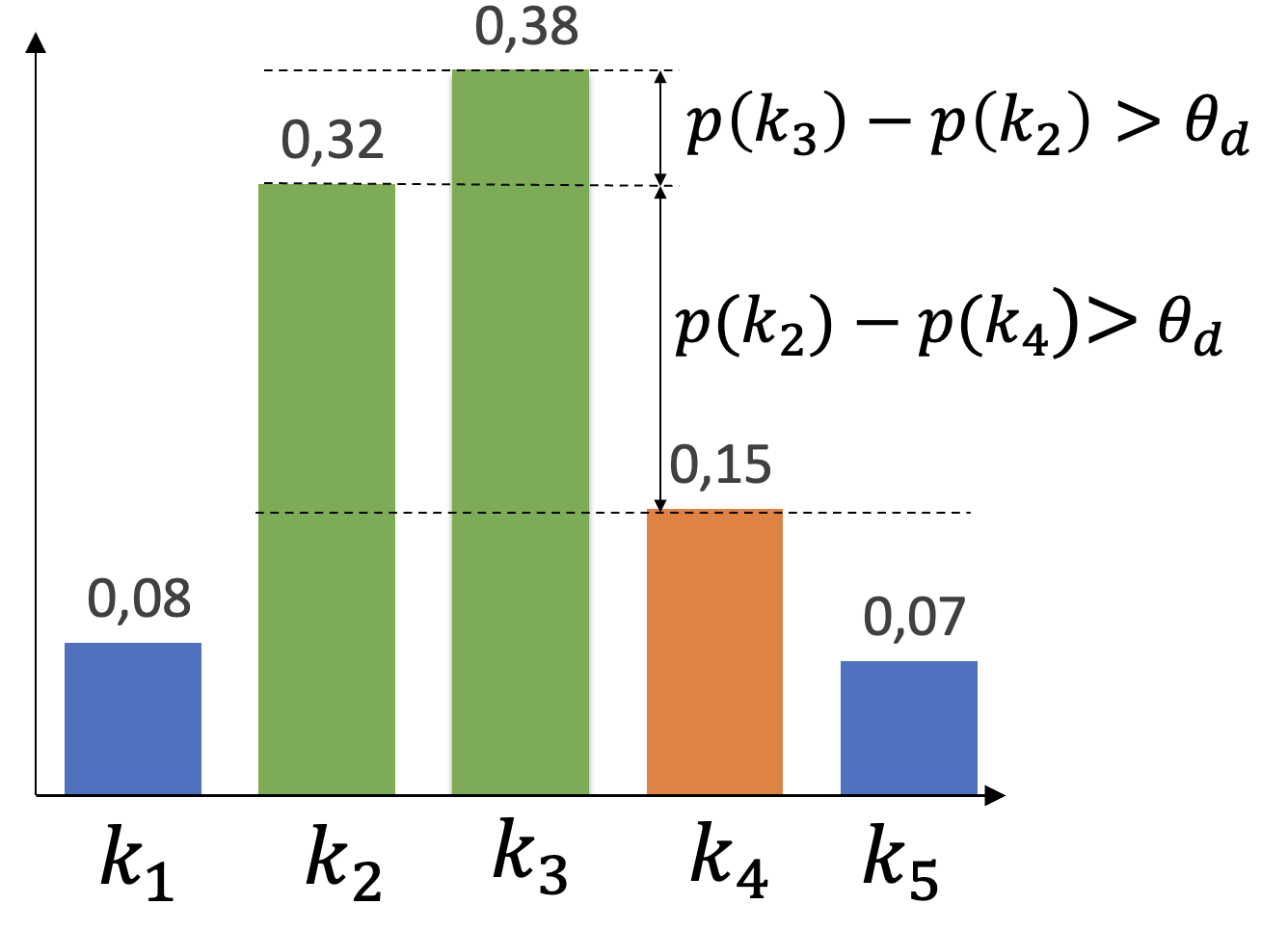}
\caption{Heuristics to select multiple labels from a probability vector}
\label{fig:heuristics}
\end{figure}

For the multiclass scenario, we assume that the network output is a categorical distribution for all the classes and use the standard supervision loss, the cross-entropy. Then, we transform the probability vector $p = [p_1, ..., p_k], \sum^k_{k=1} p_k = 1$ with three heuristics to choose the multiple winning-classes, as shows Figure \ref{fig:heuristics}. On the first method, we select the top-$k$ classes with the largest probability value $p_k$. Note that we do not considers predictions with a $s_k < 1/k$, as it occurs to $k_1, k_5$ on Figure \ref{fig:heuristics}. The second alternative is max-$\theta$, which consists in selecting all the possible classes whose $p_k$ is greater than a threshold $\theta$. Finally, the last heuristic is a dynamic $\theta_d$ threshold. We select the classes whose difference with the next class is larger than a $\theta_d$. 

On the multi-label scenario, the model outputs $\mathcal{K}$ independent Bernouilli distributions, generating binary probabilities $p = [p_1, ..., p_{k}]$, where we assume a detection if $p_k > 0.5$. Then, we substitute the cost function by a class-weighting Binary Cross Entropy (BCE) loss, obtaining $\mathcal{K}$ binary classifiers. One disadvantage of having independent binary classifiers is that the performance is more sensitive to the class imbalance in the dataset. To alleviate that, we use the Asymmetric (Asym) loss $\mathcal{L}_{asym}$ \cite{ben2020asymmetric} shown in Equation \ref{eq:asym_loss}. It combines the focal loss \cite{tian2019fcos} with the margin loss \cite{tan2019efficientnet} to reduce the contribution of easy negative samples and rejects mislabeled samples with a continuous gradient. 

\begin{equation}
\begin{split}
        ASL_k &= \left\lbrace\begin{array}{c}
     \log(p_k) (1 - p_k)^{\gamma_{+}}, \qquad y_k = 1 \\
     \log(1 - p_k) (p_k)^{\gamma_{-}}, \qquad y_k = 0 \end{array}\right. \\
    \mathcal{L}_{asym} &= \frac{1}{N} \sum^N \frac{w_k}{\mathcal{K}} \sum^{\mathcal{K}}_{k = 1} ASL_k
\end{split}
\label{eq:asym_loss}
\end{equation}

For each training image, $\mathbf{X}$ with $N$ total pixels, $\mathcal{L}_{asym}$ computes a different term depending on if the $y_k$ binary label indicating that the class $k$ is present or not in the pixel. We apply a weighting average $w_c$ depending on the ratio between positive and negative samples for class $k$ to avoid the class imbalance. Following the original paper \cite{ben2020asymmetric}, we set $\gamma^+ = 4$ and $\gamma^- = 1$.

\subsection{Example applications}

Given that our system provides metric information of the affordance location, has information of the camera poses and has multi-label affordance detection, we can apply it to common spatial tasks such as mapping and navigation.

\paragraph{Mapping of activity-centric zones}

We take the video sequence and sample unseen frames $\{f_1, ..., f_t\}$ during training. Following the same procedure as in the extraction of labels, we reproject the inferred semantic masks on the pixels $i,j$ to its respective 3D location ${x,y,z}$ using the camera intrinsic $K_{int}$, COLMAP pose $R^c_w$, $t^c_w$ and the scaled depth $d_{i,j}$. 

\begin{equation}
    \begin{bmatrix} x\\ y\\ z
\end{bmatrix} = d_{i,j} (R^c_w)^{-1} K_{int}^{-1} \begin{bmatrix}
i\\
j\\ 
1
\end{bmatrix} - t^c_w
\end{equation}

We accumulate in a global map the COLMAP key-points to represent the geometry and the segmented affordances regions. We do not perform any fusion on voxels or octrees, since our multi-label approach assumes that a zone can represent several predicted classes. Note that the map representation is common for all the sequences of the same environment, with the potential of linking zones across multiple episodes and learning from past interactions.

\paragraph{Task-oriented navigation}

Finally, we introduce a task-oriented navigation experiment to show the relevance of the map representation. We use the COLMAP key-points to build an occupancy grid with the available free space. Then, the agent is initialized in a random localization and asked to navigate to perform a certain action. Once it selects the location from the semantic-metric representation, the agent decides the path planning using a A$^*$ search with the Euclidean distance on the free space. We  use the point cloud from COLMAP to create an occupancy grid.

\begin{table*}[ht]
\centering
\resizebox{\textwidth}{!}{%
\begin{tabular}{l|cccllllllllllllllllll|}
\cline{2-22}
\multicolumn{1}{c|}{} &
  \rotatebox[origin=c]{90}{Take} &
  \rotatebox[origin=c]{90}{Put} &
  \rotatebox[origin=c]{90}{Open} &
  \rotatebox[origin=c]{90}{Close} &
  \rotatebox[origin=c]{90}{Wash} &
  \rotatebox[origin=c]{90}{Cut} &
  \rotatebox[origin=c]{90}{Mix} &
  \rotatebox[origin=c]{90}{Pour} &
  \rotatebox[origin=c]{90}{Throw} &
  \rotatebox[origin=c]{90}{Move} &
  \rotatebox[origin=c]{90}{Remove} &
  \rotatebox[origin=c]{90}{Dry} &
  \rotatebox[origin=c]{90}{Turn-on} &
  \rotatebox[origin=c]{90}{Turn} &
  \rotatebox[origin=c]{90}{Shake} &
  \rotatebox[origin=c]{90}{Turn-off} &
  \rotatebox[origin=c]{90}{Peel} &
  \rotatebox[origin=c]{90}{Adjust} &
  \rotatebox[origin=c]{90}{Empty} &
  \multicolumn{1}{l|}{\rotatebox[origin=c]{90}{Scoop}} &
  mIoU \\ \hline
  \multicolumn{1}{|l|}{} &
  \cellcolor[HTML]{804080} &
  \cellcolor[HTML]{F423E8} &
  \cellcolor[HTML]{464646} &
  \cellcolor[HTML]{66669C} &
  \cellcolor[HTML]{98FB98} &
  \cellcolor[HTML]{4682B4} & 
  \cellcolor[HTML]{FAAA1E} &
  \cellcolor[HTML]{DCDC00} &
  \cellcolor[HTML]{6B8E23} &
  \cellcolor[HTML]{BE9999} & 
  \cellcolor[HTML]{999999} & 
  \cellcolor[HTML]{DC143C} &
  \cellcolor[HTML]{FF0000} &
  \cellcolor[HTML]{00008C} &
  \cellcolor[HTML]{000046} &
  \cellcolor[HTML]{643C64} &
  \cellcolor[HTML]{CBCEFB} &
  \cellcolor[HTML]{04FFD8} &
  \cellcolor[HTML]{0000E6} &
  \cellcolor[HTML]{770B20} &
  \cellcolor[HTML]{FFFFFF} \\
  \hline
\multicolumn{1}{|l|}{GIH \cite{nagarajan2019grounded}}    & 22.1 & 21.9 & 13.8 & 10.8 & 16.3 & 25.8 & 21.2 & 23.7 & 14.0 & 16.9 & 17.2 & 12.8 & 10.3 & 20.5 & 16.6 & 10.8 & 26.1 & 9.5 & 13.9& \multicolumn{1}{l|}{25.8} & 17.5 \\ \hline

\multicolumn{1}{|l|}{Mask R-CNN \cite{he2017mask}}  & \textbf{37.7} & \textbf{36.9} & \textbf{47.1} & \textbf{43.9} & \textbf{51.5} & 41.4 &\textbf{ 46.4} & 38.1 & 43.6 & \textbf{42.9} & 38.4 & 13.1 & \textbf{52.5} & 43.7 & 30.8 & \textbf{50.9} & 35.3 & 47.1 & 33.6 & \multicolumn{1}{l|}{26.0} & 40.1 \\ \hline

\multicolumn{1}{|l|}{U-Net dyn-$\theta$\cite{ronneberger2015u}} & 0.1 & 0.7 & 5.4 & 11.9 & 22.4 & 17.1 & 22.2 & 17.3 & 11.3 & 15.9 & 21.0 & 4.5 & 14.8 & 21.1 & 16.3 & 18.4 & 12.9 & 20.6 & 0.5 & \multicolumn{1}{l|}{9.6} & 13.2 \\
\multicolumn{1}{|l|}{(ours) U-Net BCE}  & 22.3 & 22.5 & 30.9 & 24.0 & 30.2 & 23.7 & 21.1 & 17.7 & 17.1 & 23.4 & 18.5 & 8.7 & 27.3 & 22.4 & 13.2 & 23.8 & 16.2 & 22.2 & 19.0 & \multicolumn{1}{l|}{13.1} & 20.9 \\
\multicolumn{1}{|l|}{(ours) U-Net Asym} & 14.3 & 13.7 & 13.8 & 14.7 & 21.3 & 17.9 & 18.3 & 18.7 & 32.5 & 15.7 & 15.6 & 16.6 & 15.2 & 18.9 & 22.2 & 19.5 & 19.5 & 24.3 & 5.7 & \multicolumn{1}{l|}{15.7} & 17.7 \\ \hline

\multicolumn{1}{|l|}{FPN dyn-$\theta$ \cite{kirillov2019panoptic}} & 2.4 & 2.4 & 5.6 & 10.2 & 21.7 & 13.2 & 17.7 & 17.0 & 11.5 & 13.6 & 20.0 & 4.6 & 13.8 & 22.6 & 12.5 & 15.5 & 14.4 & 17.3 & 0.8 & \multicolumn{1}{l|}{9.7 } & 12.9 \\
\multicolumn{1}{|l|}{(ours) FPN BCE}  & 25.7 & 25.9 & 33.3 & 26.7 & 33.2 & 22.4 & 21.8 & 15.4 & 18.5 & 23.9 & 21.4 & 7.4 & 3.8 & 20.8 & 13.3 & 28.4 & 13.6 & 24.5 & 23.5 & \multicolumn{1}{l|}{11.6} & 22.2 \\
\multicolumn{1}{|l|}{(ours) FPN Asym} & 36.3  & 34.7  & 46.1 & 42.0 & 46.8 & 42.7 & 42.2 & 37.5 & 43.3 & 41.7 & \textbf{39.6} & 21.3 & 47.4 & 43.7 & 34.3 & 45.0 & 33.8 & 46.3 & \textbf{38.0} & \multicolumn{1}{l|}{33.2} & 39.8 \\ \hline

\multicolumn{1}{|l|}{Deep-Lab v3 dyn-$\theta$ \cite{chen2017deeplab}} & 10.1 & 11.0 & 15.4 & 17.3 & 19.1 & 19.4 & 25.2 & 19.1 & 14.7 & 17.9 & 17.1 & 9.2 & 20.4 & 31.9 & 25.3 & 26.5 & 24.3 & 31.7 & 18.0 & \multicolumn{1}{l|}{18.1} & 19.5 \\ 
\multicolumn{1}{|l|}{(ours) Deep-Lab v3 BCE}  & 33.3 & 34.2 & 44.1 & 37.6 & 43.1 & 32.0 & 30.9 & 26.2 & 28.9 & 33.7 & 27.6 & 13.2 & 41.6 & 27.6 & 22.2 & 39.1 & 22.3 & 35.1 & 32.3 & \multicolumn{1}{l|}{20.7} & 31.3 \\
\multicolumn{1}{|l|}{(ours) Deep-Lab v3 Asym} & 31.6 & 32.9 & 37.3 & 37.8 & 44.5 & \textbf{43.9} & 45.0 & \textbf{41.8} & \textbf{53.4} & 42.3 & 39.4 & \textbf{33.1} & 45.5 & \textbf{52.2} & \textbf{44.0} & 46.7 & \textbf{43.5} & \textbf{51.1} & 32.3 & \multicolumn{1}{l|}{\textbf{46.6}} & \textbf{42.3} \\ \hline
\end{tabular}%
}
\caption{Class-wise IoU scores on easy-EPIC Aff test set. All scores are in [$\%$].}
\label{tab:class_wise_scores}
\end{table*}

\begin{table}[htbp]
\centering
\resizebox{\columnwidth}{!}{%
\begin{tabular}{|cccc|cccc|}
\hline
\multicolumn{1}{|c|}{} & KLD $\downarrow$ & SIM $\uparrow$ & AUC-J $\uparrow$ & mIoU $\uparrow$ & F1-Score $\uparrow$ & mAP $\uparrow$ & AP50 $\uparrow$ \\ \hline \hline
\multicolumn{1}{|c|}{GIH  \cite{nagarajan2019grounded}} & 2.381 & 0.116 & 0.511 & 17.5 & 29.4 & 14.2 & 15.5 \\ \hline

\multicolumn{1}{|c|}{Mask-RCNN \cite{he2017mask}} & 1.365 & 0.150 & 0.841 & 40.1 & 56.5 &\textbf{ 59.3} & \textbf{62.6} \\ \hline \hline

\multicolumn{1}{|c|}{U-Net \cite{ronneberger2015u} top-$\mathcal{K}$} & 2.532 & 0.341 & 0.830 & 9.5 & 17.4 & 22.0 & 30.5 \\
\multicolumn{1}{|c|}{U-Net \cite{ronneberger2015u} max-$\theta$} & 2.532 & 0.341 & 0.830 & 13.2 & 23.6 & 22.0 & 30.5 \\
\multicolumn{1}{|c|}{U-Net \cite{ronneberger2015u} dyn-$\theta$} & 2.532 & 0.341 & 0.830 & 13.2 & 23.7 & 22.0 & 30.5  \\ \hline
\multicolumn{1}{|c|}{(ours) U-Net + BCE} & 2.718 & 0.304 & 0.949 & 20.9 & 34.2 & 48.2 & 44.7 \\
\multicolumn{1}{|c|}{(ours) U-Net + Asym} & 0.783 & 0.665 & 0.857 & 17.7 & 29.9 & 15.6 & 32.3 \\ \hline \hline

\multicolumn{1}{|c|}{FPN \cite{kirillov2019panoptic} top-$\mathcal{K}$} & 2.229 & 0.362 & 0.812 & 8.9 & 15.6 & 18.9 & 24.7 \\
\multicolumn{1}{|c|}{FPN \cite{kirillov2019panoptic} max-$\theta$} & 2.229 & 0.362 & 0.812 & 12.4 & 21.8 & 18.9 & 24.7 \\
\multicolumn{1}{|c|}{FPN \cite{kirillov2019panoptic} dyn-$\theta$} & 2.229 & 0.362 & 0.812 & 12.9 & 23.6 & 18.9 & 24.7 \\ \hline 
\multicolumn{1}{|c|}{(ours) FPN + BCE} & 1.613 & 0.365 & 0.955 & 22.2 & 35.7 & 48.7 & 44.5 \\
\multicolumn{1}{|c|}{(ours) FPN + Asym} & 0.789 & 0.546 & 0.956 & 39.8 & 56.8 & 44.1 & 59.3 \\  \hline \hline

\multicolumn{1}{|c|}{DeepLab-v3 \cite{chen2017deeplab} top-$\mathcal{K}$} & 4.947 & 0.192 & 0.911 & 18.9 & 31.9 & 35.0 & 40.9 \\
\multicolumn{1}{|c|}{DeepLab-v3 \cite{chen2017deeplab} max-$\theta$} & 4.947 & 0.192 & 0.911 & 19.2 & 32.3 & 35.0 & 40.9 \\
\multicolumn{1}{|c|}{DeepLab-v3 \cite{chen2017deeplab} dyn-$\theta$} & 4.947 & 0.192 & 0.911 & 19.5 & 32.7 & 35.0 & 40.9 \\ \hline 
\multicolumn{1}{|c|}{(ours) DeepLab-v3 + BCE} & 1.276 & 0.179 & 0.964 & 31.3 & 47.2 & 58.6 & 56.2 \\ 
\multicolumn{1}{|c|}{(ours) DeepLab-v3 + Asym} & \textbf{0.603} & \textbf{0.668} & \textbf{0.965} & \textbf{42.3} & \textbf{60.1} & 43.6 & 58.5 \\ \hline \hline

\end{tabular}%
}
\caption{Affordance multi-label segmentation on easy-EPIC Aff test set (20 classes). Note that except the mIoU and the F1-Score, the rest of the metrics are common for the three versions of the multi-class segmentation models.}
\label{tab:baselines_20_classes}
\end{table}

\begin{table}[htbp]
\centering
\resizebox{\columnwidth}{!}{%
\begin{tabular}{|cccccccc|}
\hline
\multicolumn{1}{|c|}{} & KLD $\downarrow$ & SIM $\uparrow$ & \multicolumn{1}{c|}{AUC-J $\uparrow$}  & mIoU $\uparrow$ & F1-Score $\uparrow$ & mAP $\uparrow$ & AP50 $\uparrow$ \\ \hline
\multicolumn{1}{|c|}{Mask-RCNN} & 2.287 & 0.211 & \multicolumn{1}{c|}{0.756} & 17.1 & 27.3 & \textbf{40.1} & \textbf{46.7} \\ 
\multicolumn{1}{|c|}{(ours) U-Net Asym} & 1.104 & 0.320 & \multicolumn{1}{c|}{0.657} & 12.9 & 24.8 & 11.2 & 17.9 \\
\multicolumn{1}{|c|}{(ours) FPN Asym} & 0.530 & \textbf{0.673} & \multicolumn{1}{c|}{0.921} & 28.1 & 42.9 & 24.8 & 43.4 \\
\multicolumn{1}{|c|}{(ours) DeepLab-v3 Asym} &\textbf{ 0.520} & 0.670 & \multicolumn{1}{c|}{\textbf{0.931}} & \textbf{31.1 }& \textbf{46.5} & 27.4 & 43.9 \\ \hline
\end{tabular}%
}
\caption{Affordance multi-label segmentation on complex-EPIC Aff test set (43 classes).}
\label{tab:baselines_43_classes}
\end{table}

\section{Experiments}

\subsection{Models and metrics}

In our experiments, we modify three popular semantic segmentation architectures \cite{kirillov2019panoptic, ronneberger2015u, chen2017deeplab} and compare them with a instance segmentation model \cite{he2017mask} plus an interaction hotspots model \cite{nagarajan2019grounded}.
\begin{itemize}
    \item Grounded Interaction Hotspots (GIH) \cite{nagarajan2019grounded}: We use the weights on its EPIC-Kitchen trained version to extract predictions from our images. To reduce the gap, we crop our scenes to represent a single object and compare for the same number of affordable actions $\mathcal{K}$ in the easy-EPIC Aff dataset.
    
    \item Mask-RCNN \cite{he2017mask}: we assume an overlapping in the bounding boxes between two different instances. We do not consider the amodal Mask-RCNN versions \cite{qi2019amodal, mohan2022amodal} which treat differently visible and occlusion masks, since our affordance classes $\mathcal{K}$ are not ranked in order.
    
    \item Semantic segmentation architectures. We compare the performance of UNet \cite{ronneberger2015u}, Feature Pyramid Networks (FPN) \cite{kirillov2019panoptic, lin2017feature} and DeepLab v-3 \cite{chen2017deeplab}.
\end{itemize}

We train the segmentation models with an input resolution of 232 $\times$ 348 for 100 $k$ iterations usign Adam as optimizer with weight decay of $10^{-4}$, batch size of 8 and a initial learning rate of $10^{-4}$, using a polynomial decay up to $10^{-6}$. We apply random crop, color jitter, resize and flipping as data augmentation. In the same way, we train Mask-RCNN SGD and $10^{-2}$ as initial learning rate. We use a Resnet-50 backbone pre-trained on Imagenet for all the models in order to perform a fair comparative.

Following the evaluation of Nagarajan et al, \cite{nagarajan2019grounded}, we report the Kullback-Leibler Divergence (KLD) \cite{fang2018demo2vec}, the Similarity metric (SIM) and the Area Under the Curve (AUC-J) \cite{bylinskii2018different,judd2009learning} which provide different metrics for the mismatch of the distribution of heatmaps or affordance regions considering the predictive probability. We also report metrics from segmentation literature, such as the mean Intersection over the Union (mIoU) and the F1-Score to measure the performance of the semantic segmentation, and the Average Precision (AP) AP-50 and mAP to report the performance of the detection metrics.

\begin{figure}[tbp]
\centering
\resizebox{\columnwidth}{!}{%
\includegraphics[width=0.66\columnwidth]{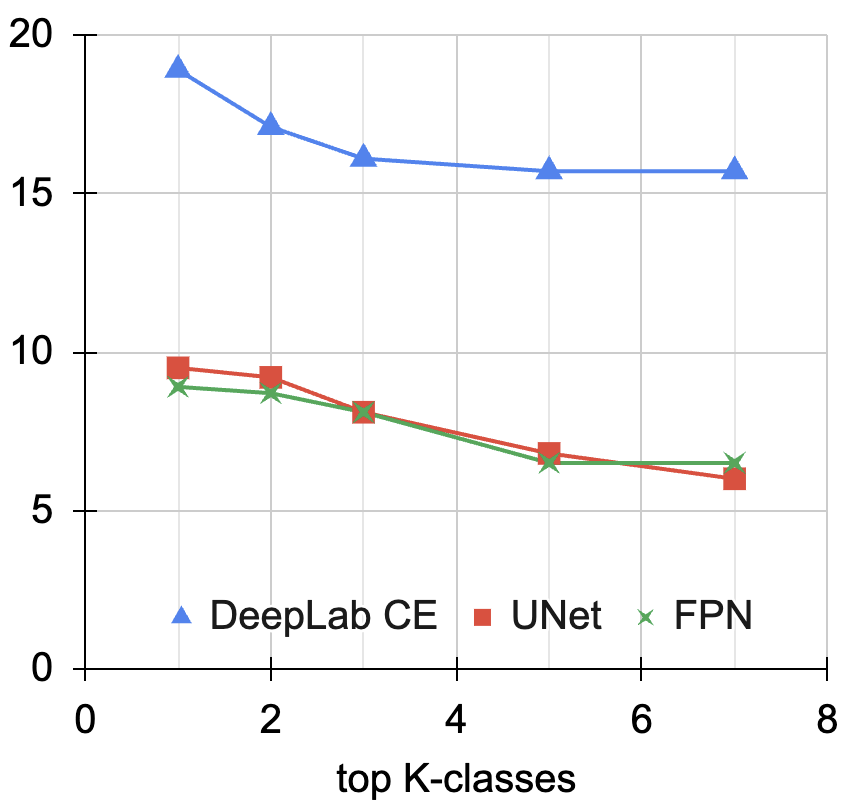} 
\includegraphics[width=0.66\columnwidth]{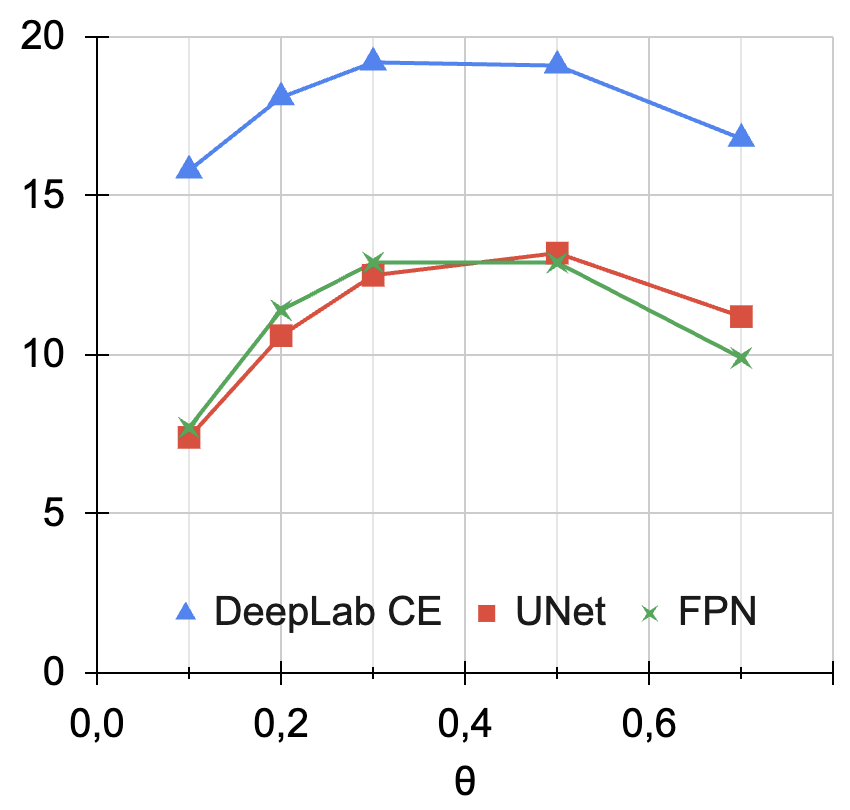} 
\includegraphics[width=0.66\columnwidth]{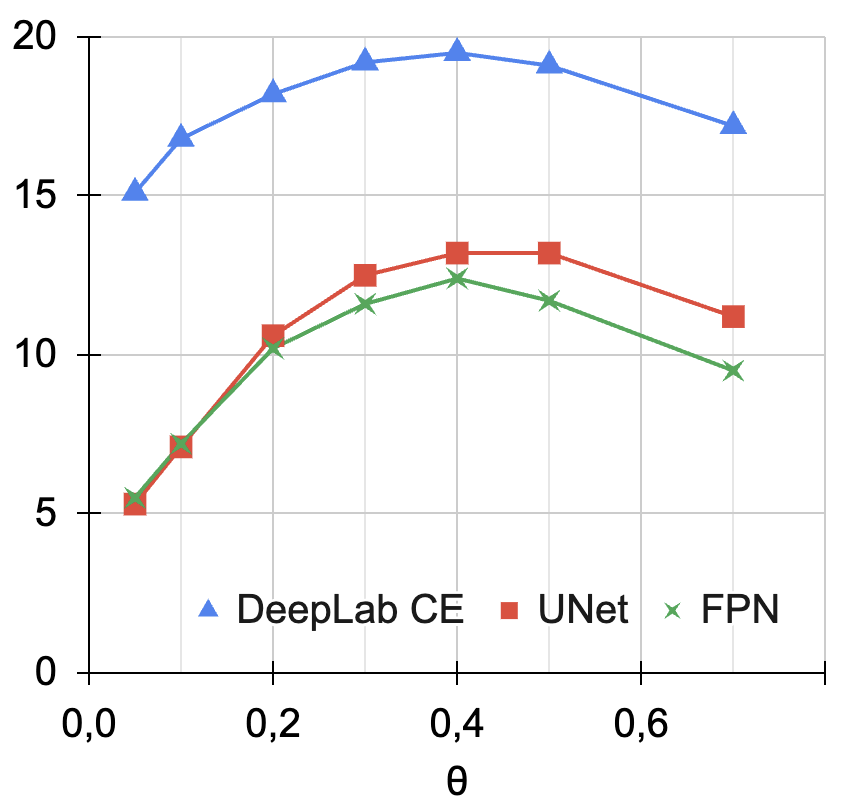}
}
\caption{Evolution of the mIoU for different heuristics to select multiple winning classes from a multi-class probability vector. Left: top-$\mathcal{K}$. Center: max-$\theta$. Right: dyn-$\theta$.}
\label{fig:ce}
\end{figure}

\begin{figure*}[t]
\centering
\includegraphics[width=0.66\columnwidth]{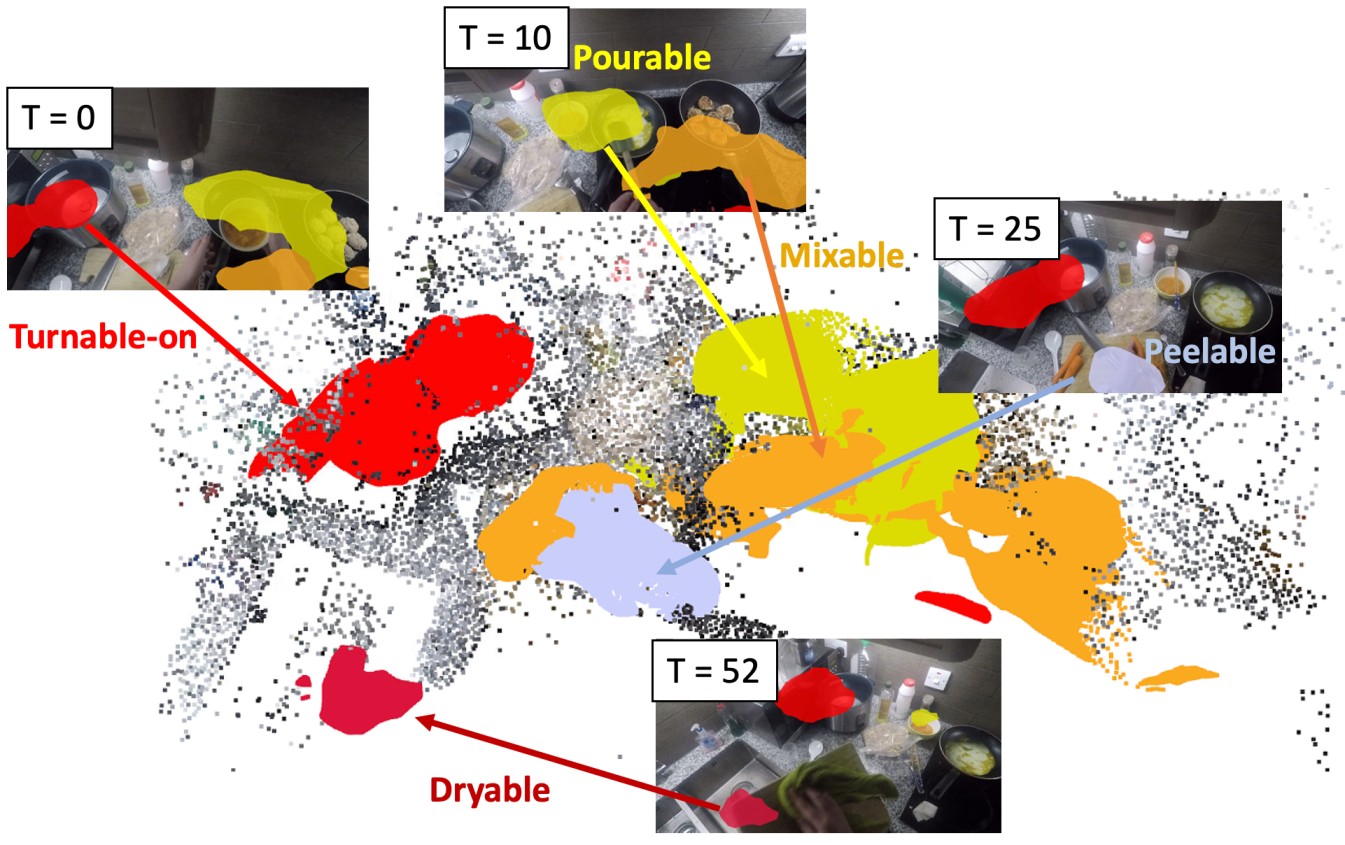} 
\includegraphics[width=0.66\columnwidth]{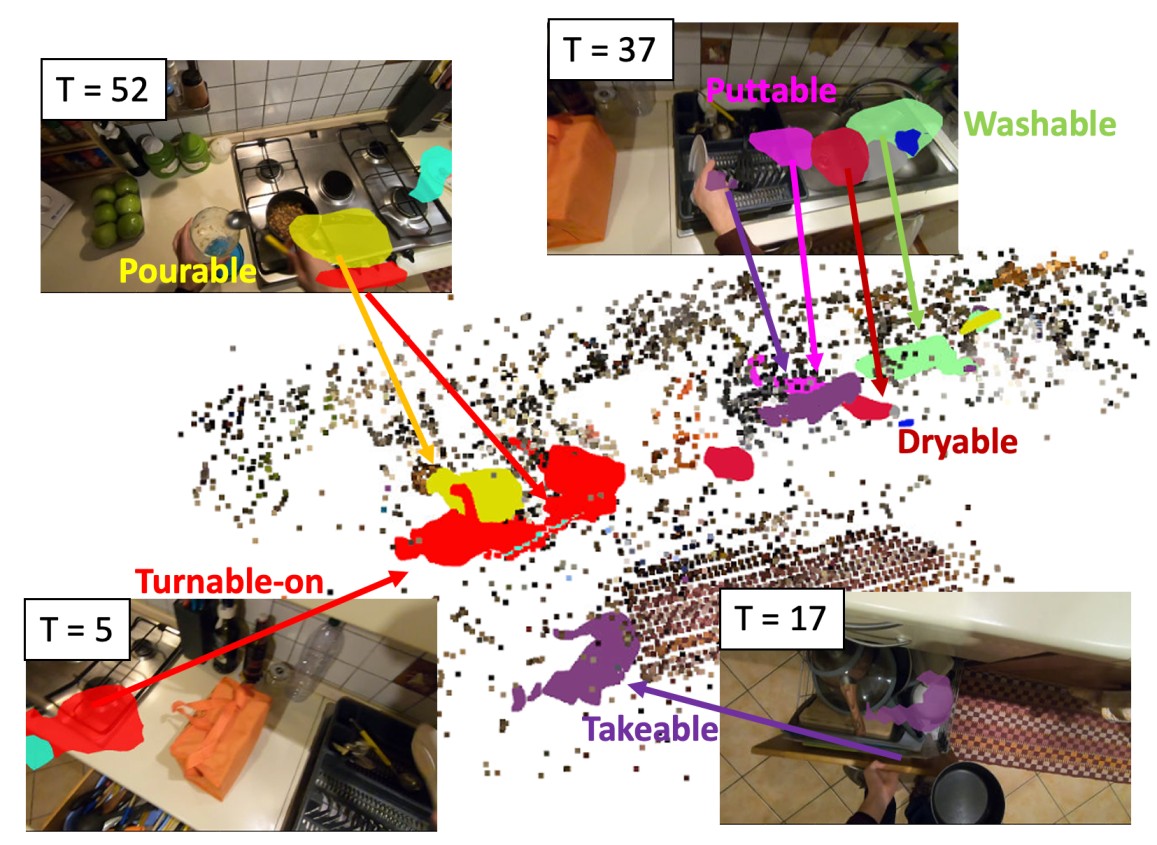} 
\includegraphics[width=0.66\columnwidth]{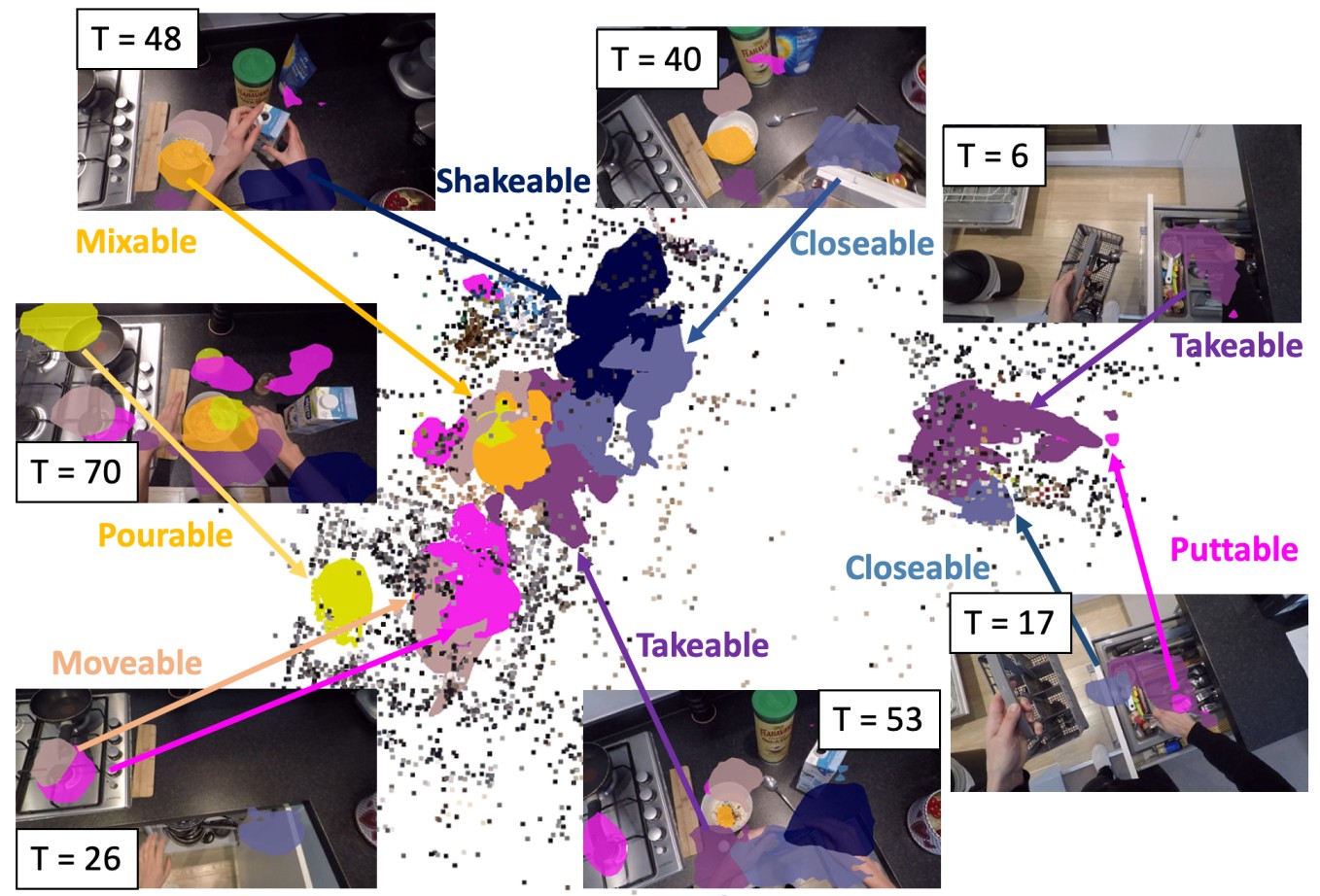}
\caption{Spatial distribution of the detected multi-label affordances for multiple time-steps}
\label{fig:aff_map}
\end{figure*}

\begin{figure}[t!]
\centering
\includegraphics[width=0.99\columnwidth]{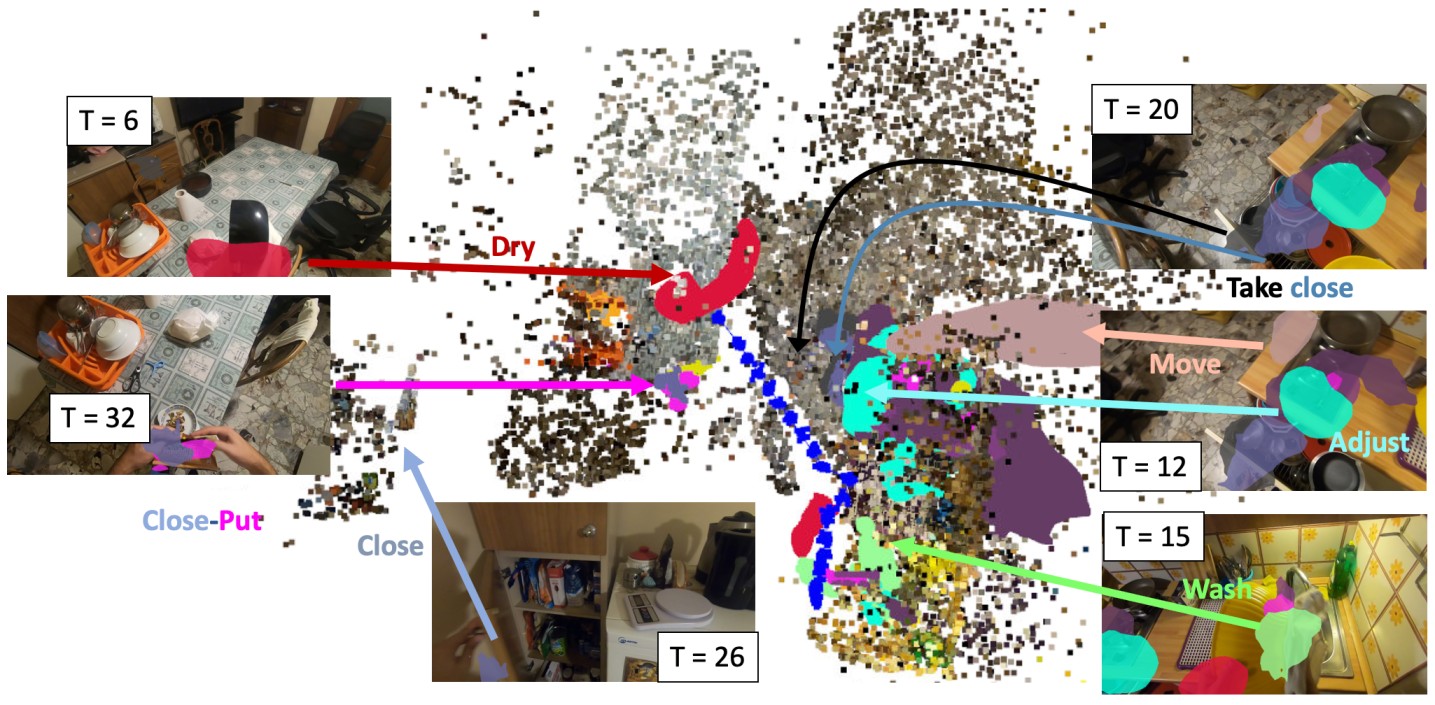}
\caption{Goal oriented path-planning. In the example, at t=36 we indicate the user the trajectory from the sink to the place where it used to dry the crockery. The blue points represents the steps of the path planning}
\label{fig:final aplication}
\end{figure}

\subsection{Quantitative results}

We compare the performance of different popular architectures on the multi-label affordance segmentation task in Table \ref{tab:baselines_20_classes} on the easy-EPIC Aff dataset. DeepLab-v3 trained with the Asymmetric loss obtains the best performance on the segmentation and saliency metrics (42.3~$\%$ mIoU 60.1~$\%$ F-1 score, 0.603 KLD, 0.668 SIM, 0.965 AUC-J). Since the backbone of the three semantic segmentation models is the same (Resnet-50), the different results are due to the configuration of the different decoders. In the dataset, since the labels represent interaction hotspots, they are not aligned with the borders of the objects and represent a more high-level zone. Thus, the atrous convolution of DeepLab enables to enlarge of the filter's field of view and better captures these regions. We show the per-class segmentation performance in terms of the IoU in Table \ref{tab:class_wise_scores}. Comparing with the apparition frequency of the classes in the dataset shown in Figure \ref{fig:distrib_classes}, Mask R-CNN fails at the low-represented classes since is not trained with the Asymmetric loss. However, it is the best architecture on the detection metrics (59.3~$\%$ mAP, 62.6~$\%$ AP50). Compared with previous works, the pre-trained version of \cite{nagarajan2019grounded} achieves intermediate results on the segmentation metrics (17.5 $\%$ mIoU, 29.4 $\%$ F-1 score) but low on the AP scores.

The results in Figure \ref{fig:ce} show the impact of the hyper-parameters when adapting the multi-class models. The top-$\mathcal{K}=1$ represents the classical multi-label case. The results show how its performance is far from the multi-label versions, supporting the need for specific architectural changes for this scenario. In Figure \ref{fig:ce} left, when we increase the number of winning classes, the performance decreases by introducing too many false positives. The other two heuristics achieve better performance since they better reject these outliers. For example, the dyn-$\theta$ adapts dynamically to the probability distribution shape, obtaining higher mIoU and F-1 in the three cases (see Table \ref{tab:baselines_20_classes}). 
Finally, we appreciate similar results on the complex-EPIC Aff, shown in Table \ref{tab:baselines_43_classes}. In this case, the overall performance decreases due to the higher number of classes and its imbalance.


\subsection{Mapping: metric distribution of affordances}

We show on Figure \ref{fig:aff_map} qualitative results of the multi-label interaction hotspots from affordances predicted by the DeepLab-v3 Asym model. Our perception model is consistent from different view-points. For example, the microwave of the left-map in Figure \ref{fig:aff_map} is detected as \textit{turn-on} both at $t=0, 25, 52$. The qualitative results clearly motivate our multi-label approach: the milkshake on the right map affords \textit{mixing}, \textit{pouring} and \textit{taking}, or the sink in the center map affords \textit{drying} and \textit{washing}. The metric conception of our approach is also relevant, since it reflects the interactions hotspots rather than highlight the complete object (for example \textit{grasping} a pan only with the handle).

\subsection{Task-oriented navigation}

Finally, we use the spatial localization of the affordances to show a proof-of-concept "task-oriented" navigation. As we illustrate on Figure \ref{fig:final aplication}, we guide the agent according to the action possibilities that the environment offers to him. 
Therefore, we can ask our system to \emph{perform} certain action, meaning to \emph{go to where the object and affordance are available}. The A$^*$ indicates to the agent the shortest path from its current location to the position where it took the action in the past. For example, this could guide a visually impaired person with an assistant device \cite{guerreiro2020virtual}.

\section{Limitations}

Our current approach presents several limitations. At the dataset extraction, we assume that the interaction occurs in the intersection between the object and the hand bounding-boxes, thus it depends on the bounding-box aligned to the actual object. This could be mitigated with a detection model for grasping points, but we wanted a simpler version for our prototype as a more convoluted approach might introduce further biases, difficult to detect. Also, the camera poses from COLMAP can be distorted by noisy-frames or dynamic objects non-suppressed by the mask. Furthermore, a real-time mapping system would require a SLAM system such as ORB-SLAM \cite{mur2015orb} which might reduce the accuracy of COLMAP. Similarly, our dataset is fully based on Kitchen sequences and it does not incorporate another environments introducing important dataset bias in the trained models. However, our automatic labeling pipeline could be easily used to extend the dataset in other scenarios.

\section{Conclusions}

We introduced a novel multi-label, metric and spatial-oriented perception of affordances. First, we present a method for extracting grounded affordances labels based on egocentric interaction videos through a common metric representation of all the past interactions in a common reference. We use this pipeline to build the most complete affordance dataset based on the classic EPIC-Kitchen dataset. This constitutes EPIC-Aff, the largest semantic segmentation dataset of affordances grounded on the human interactions. We also motivate a method for grounded affordance detection with pixel precission using multi-label predictors, which enhances the perception and the representation of the environment. Furthermore, we show that the metric representation obtained can be used to build detailled affordance maps and to guide the user to perform task-oriented navigation tasks.

\section*{Acknowledgements}
This work was supported by the Spanish Government (PID2021-125209OB-I00, TED2021-129410B-I00 and TED2021-131150B-I00) and the Aragon Government (DGA-T45\textunderscore23R).

{\small
\bibliographystyle{ieee_fullname}
\bibliography{iccv2023AuthorKit/mybib}
}

\end{document}